\documentclass[conference, a4paper]{IEEEtran}
\IEEEoverridecommandlockouts
\usepackage{geometry}
\usepackage{cite}
\usepackage{amsmath,amssymb,amsfonts}
\usepackage{algorithmic}
\usepackage{graphicx}
\usepackage{textcomp}
\usepackage{xcolor}
\usepackage{graphicx}
\usepackage{subfigure}
\usepackage{CJKutf8}
\usepackage{algorithm}
\usepackage{algorithmic}

\newtheorem{assumption}{Assumption}
\newenvironment{sloppypar*}
{\sloppy\ignorespaces}
{\par}
\def\BibTeX{{\rm B\kern-.05em{\sc i\kern-.025em b}\kern-.08em
    T\kern-.1667em\lower.7ex\hbox{E}\kern-.125emX}}

\begin{document}\sloppy

\title{Semi-Supervised Federated Learning with non-IID Data: Algorithm and System Design}

\author{\IEEEauthorblockN{Zhe Zhang\textsuperscript{1,2}, Shiyao Ma\textsuperscript{3}, Jiangtian Nie\textsuperscript{4}, Yi Wu\textsuperscript{1,2,*}, Qiang Yan\textsuperscript{5}, Xiaoke Xu\textsuperscript{3}, Dusit Niyato\textsuperscript{4}, \textit{Fellow, IEEE}}
\IEEEauthorblockA{\textit{\textsuperscript{1}School of Data Science and Technology, Heilongjiang University, Harbin 150080, China} \\
\textit{\textsuperscript{2}Cryptology and Network Security Laboratory of Heilongjiang University, Harbin 150080, China} \\
\textit{\textsuperscript{3}College of Information and Communication Engineering, Dalian Minzu University} \\
\textit{\textsuperscript{4}School of Computer Science and Engineering, Nanyang Technological University}\\
\textit{\textsuperscript{5}WeBank Co. Ltd., China}\\
\textsuperscript{*}1995050@hlju.edu.cn}}

\maketitle

\begin{abstract}
Federated Learning (FL) allows edge devices (or \emph{clients}) to keep data locally while simultaneously training a shared high-quality global model. 
However, current research is generally based on an assumption that the training data of local clients have ground-truth. 
Furthermore, FL faces the challenge of statistical heterogeneity, i.e., the distribution of the client's local training data is non-independent identically distributed (non-IID).
In this paper, we present a robust semi-supervised FL system design, where the system aims to solve the problem of data availability and non-IID in FL.
In particular, this paper focuses on studying the labels-at-server scenario where there is only a limited amount of labeled data on the server and only unlabeled data on the clients.
In our system design, we propose a novel method to tackle the problems, which we refer to as Federated Mixing ($\mathrm{FedMix}$). $\mathrm{FedMix}$ improves the naive combination of FL and semi-supervised learning methods and designs parameter decomposition strategies for disjointed learning of labeled, unlabeled data, and global models.
To alleviate the non-IID problem, we propose a novel aggregation rule based on the frequency of the client's participation in training, namely the $\mathrm{FedFreq}$ aggregation algorithm, which can adjust the weight of the corresponding local model according to this frequency.
Extensive evaluations conducted on CIFAR-10 dataset show that the performance of our proposed method is significantly better than those of the current baseline. It is worth noting that our system is robust to different non-IID levels of client data.
\end{abstract}


\section{Introduction}
Federated Learning (FL) \cite{mcmahan2017communication, 8832210} is a distributed machine learning paradigm that allows multiple edge devices (or \emph{clients}) to cooperatively train a shared global model \cite{liu2020privacy,lim2020information,liu2020deep}.
The most obvious difference between FL and traditional distributed machine learning is that clients can privately access local training data without sharing data with cloud centers \cite{li2018federated,liu2020federated,kang2020reliable}.
However, the current mainstream work is based on an unrealistic assumption: \emph{the training data of local clients have ground-truth} \cite{liu2020rc}.
In our daily lives, it is not common for each client to have rich labeled data. For example, in the early stage of COVID-19 epidemic, community hospitals without enough labeled data may not be able to train a high-precision pathophoresis prediction model. 
On the other hand, in most cases, putting together a properly labeled dataset for a given FL task is a time-consuming, expensive, and complicated endeavor \cite{liu2020rc}. 
Therefore, it is challenging to train a high-quality global model in the real scenario of a lack of labeled data.

In the face of the above challenges, recent works \cite{liu2020rc,long2020fedsemi,jin2020towards,jeong2020federated,zhu2021semi,wang2020graphfl} study how to design a semi-supervised FL (SSFL) system that can efficiently integrate semi-supervised learning into FL techniques.
For example, Jeong \textit{et al.} in \cite{jeong2020federated} proposed an SSFL system with a new inter-client consistency loss to achieve this goal. 
In fact, consistency regularization technique is widely used in semi-supervised learning which keeps the same output that the same data inject two different noises \cite{samuli2017temporal,park2018adversarial}. 
Furthermore, pseudo-label methods are important for SSFL, where they mainly utilize pseudo-labels whose predicted value is higher than the confidence threshold to achieve high-precision SSFL \cite{liu2020rc,long2020fedsemi,sohn2020fixmatch}. 
However, it is worth noting that there still remain gaps when deploying SSFL in practice. 

First, traditional SSFL methods generally introduce semi-supervised techniques (such as consistency loss and pseudo-label) directly into the FL system, which ignores the implicit contribution between iterative updates of the global model. 
Previous work only focused on how to set pseudo-labels or how to decompose the parameters of labeled and unlabeled data for disjoint learning. 
In this way, the learned global model will be biased towards labeled data (supervised model) or unlabeled data (unsupervised model) instead of the global model \cite{jeong2020federated}. 
This implies that we need to observe the implicit effects between iterations of the global model at a fine-grained level.

Second, the non-independent identically distributed (non-IID) of data between clients has always been a key and challenging issue in FL. 
The reason is that there are too many differences in data distribution, features, and the number of labels between clients, which is not conducive to the convergence of the global model. 
Currently, many efforts have effectively alleviated the non-IID problem, such as $\mathrm{FedBN}$ \cite{li2020fedbn} utilized local batch normalization to alleviate the feature shift before average aggregating local models. 
However, methods such as these add additional computational and communication overhead to the server or client. 

In this paper, to address the first issue, we propose the Federated Mixing ($\mathrm{FedMix}$) algorithm, which performs parameter decomposition of disjointed learning for supervised model (learned on labeled data), unsupervised model (learned on unlabeled data), and global model. In particular, this algorithm analyzes the implicit effects between iterations of the global model in a fine-grained manner. To address the second issue, we propose a novel aggregation rule called Federated Frequency ($\mathrm{FedFreq}$), which dynamically adjusts the weight of the corresponding local model by recording the training frequency of the client to alleviate the non-IID problem. Furthermore, we introduce the Dirchlet distribution function to simulate the different non-IID level scenario in our experiment. The main contributions of this paper are as follows:
\begin{itemize}
    \item
    We present a robust Semi-supervised Federated Learning system design, where the system aims to solve the problem of data availability and non-IID in FL. In our system design, we propose $\mathrm{FedMix}$ algorithm to improve the naive combination of FL and semi-supervised learning methods. 
    \item
    We propose a novel aggregation rule called $\mathrm{FedFreq}$, which dynamically adjusts the weight of the corresponding local model by recording the training frequency of the client to alleviate the non-IID problem.
    \item
    We conduct extensive evaluations on CIFAR-10 dataset, which show that the performance of our designed system is 3\% higher than the baseline.
\end{itemize}

\section{related work}
\subsection{Semi-supervised Federated Learning}
Semi-supervised federated learning attempts to use semi-supervised learning techniques \cite{zhu2009introduction,chapelle2009semi,kingma2014semi,zhai2019s4l,mallapragada2008semiboost} to further improve the performance of the FL model in scenarios where there is unlabeled data on the client side \cite{jin2020towards}. For example, Long \textit{et al.} in \cite{long2020fedsemi} proposed a semi-supervised federated learning (SSFL) system, \textit{FedSemi}, which unifies the consistency-based semi-supervised learning model \cite{lee2013pseudo}, dual model \cite{samuli2017temporal}, and average teacher model \cite{tarvainen2017mean} to achieve SSFL. The DS-FL system \cite{9392310} was proposed to solve the communication overhead problem in SSFL. Reference \cite{zhang2020improving} proposes a method to study the distribution of non-IID data, which introduces a probability distance metric to evaluate the difference in client data distribution in SSFL. Different from the literature \cite{liu2020rc,long2020fedsemi,9392310}, in this paper, we focus on labels-at-server scenario and also solve the problem of data availability and data heterogeneity in SSFL.

\subsection{Robust Federated Learning}
If the local data set distribution of each client is inconsistent (i.e., non-IID problem) \cite{li2019convergence,zhao2018federated,sattler2019robust,briggs2020federated,li2020fedbn,wang2020optimizing,chen2020asynchronous}, the local objective loss function of the client will be inconsistent with the global objective\cite{li2021federated}. In particular, when the model of the local client is updated larger, such a difference will be more obvious. Therefore, we need to design some robust FL systems to solve the above problems. Some studies try to design a robust federated learning algorithm to solve the non-IID problem. For example, \textit{FedProx} \cite{li2018federated} limits the distance between the local model and the global model by introducing an additional $\mathcal{L}_2$ regularization term in the local target function to limiting the size of the local model update. However, this method has a disadvantage that each client needs to individually adjust the local regularization term to obtain good model performance. \textit{FedNova} \cite{wang2020tackling} improved FedAvg in the aggregation phase, which normalized and scaled the model update according to the local training batch of the client. Although previous studies have alleviated the problem of non-IID to some extent, they only evaluated the data distribution at specific non-IID levels and lacked extensive experimental verification for different non-IID scenarios. Therefore, we propose a more comprehensive data distribution and data partition strategy, i.e., we introduce the Dirichlet distribution function to simulate different non-IID levels of client data.
\section{Preliminaries}
\subsection{Federated Learning}
Federated learning solves the problem of data island on the premise of privacy protection.
In particular, the FL is a distributed machine learning framework, which requires clients to hold data locally, where these clients coordinate to train a shared global model $\omega^\ast$.
In FL, there is a server $\mathcal{S}$ and $k$ clients, each of which holds an IID or non-IID datasets $\mathcal{D}_k$.
Specifically, for a training sample $x$ on the client side, let $\ell(\omega; x)$ be the loss function at the client, where $\omega  \in {\mathbb{R}^d}$ denotes the model’s trainable parameters.
Therefore, we let $\mathcal{L}(\omega) = \mathbb{E}_{x \sim \mathcal{D}}[\ell(\omega ; x)]$ be the loss function at the server. Thus, FL needs to optimize the following objective function at the server:
\begin{equation}
	{\min _\omega }\mathcal{L}(\omega ),{\text{ where }}\mathcal{L}(\omega ): = \sum\limits_{k = 1}^K {{p_k}} {\mathcal{L}_k}(\omega ),
\end{equation}
where ${p_k\ge 0},\sum\limits_k {{p_k}}  = 1$ indicates the relative influence of $k$-th client on the global model. In FL, to minimize the above objective function, the server and clients execute the following steps:
\begin{itemize}
    \item \textbf{\textit{Step 1, Initialization:}} The server sends the initialized global model $\omega_0$ to the selected clients.
    \item \textbf{\textit{Step 2, Local training:}} The client uses the local optimizer (e.g., SGD, Adam) on the local dataset $\mathcal{D}_k$ to train the received initialization model. Then, each client uploads the local model $\omega_{t}^{k}$ to the server.
    \item \textbf{\textit{Step 3, Aggregation:}} The server collects and uses a certain algorithm (e.g., FedAvg \cite{mcmahan2017communication}) to aggregate the model updates uploaded by these clients to obtain a new global model, i.e., $\omega_{t+1}$ = $\omega_{t}$ +  $\sum\limits_{k = 1}^K \dfrac{{D}_k}{D} \omega_{t}^{k}$. 
Then, the server sends the updated global model $\omega_{t+1}$ to all selected clients.
\end{itemize}

Note that FL repeats the above steps until the global model converges.

\subsection{Semi-supervised Learning}
In the real world, (e.g., financial and medical fields), unlabeled data is easy to gain, while labeled data is often difficult to obtain. 
Meanwhile, annotating data requires a lot of manpower and material resources. To this end, the researchers proposed a machine learning paradigm namely Semi-supervised Learning \cite{sajjadi2016regularization,xie2020unsupervised} to learn a high-precision model on a mixed dataset (part of the data is labeled, and some of the data is unlabeled). 
Thus, in recent years, semi-supervised learning has become a hot research direction in the field of deep learning. 
In this section, we introduce a basic assumption and two methods of semi-supervised learning.

\begin{assumption}
\textit{In machine learning, there is a basic assumption that if the feature of two unlabeled samples $u_1$ and $u_2$ are similar, the corresponding model prediction results $y_1$ and $y_2$ are the same \cite{yang2021survey}, i.e., $f({u_1}) = f({u_2})$, where $f( \cdot )$ is the prediction function.}
\end{assumption}

According to the above assumption, we adopt two common semi-supervised learning methods as follows:

\textbf{Consistency Regularization}: The main idea of this method is that the model prediction results should be the same whether noise is added or not on an unlabeled training sample \cite{samuli2017temporal,sajjadi2016regularization}. 
We generally use data augmentation (such as image flipping and shifting) methods to add noise to increase the diversity of the dataset. 
Specifically, for an unlabeled sample $u_i$ in unlabeled dataset $\mathbf{u}=\{u_i\}_{i=1}^{m}$ and its perturbation form $\hat{u_i}$, our goal is to minimize the distance $d(f_\theta (u_i),f_\theta (\hat{u_i}))$, where $f_\theta (u_i)$ is the output of sample $u_i$ on model $\theta$.
The common distance measurement method is Kullback-Leiber (KL) divergence.
Thus, we can calculate the consistency loss as follows:

\begin{equation}
d (f_\theta; \mathbf{u})_{KL} = \dfrac{1}{m} \sum\limits_{i = 1}^m f_\theta (u_i) log\dfrac{f_\theta (u_i)}{f_\theta (\hat{u_i})},
\end{equation}
where $m$ is the total number of unlabeled samples and $f_\theta (u_i)$ indicates the model output of unlabeled sample $u_i$.

\textbf{Pseudo-label}: 
The pseudo-label method \cite{lee2013pseudo} is to utilize some labeled samples to train a model that to set the pseudo labels for unlabeled samples.
Previous work generally used sharpening \cite{berthelot2019mixmatch} and argmax \cite{lee2013pseudo} methods to set pseudo-label, where the former make the distribution of model output extreme of unlabeled samples and the latter change the model output to one-hot of unlabeled samples.

\begin{figure*}[!t]
	\centering
	\includegraphics[width=0.7\linewidth]{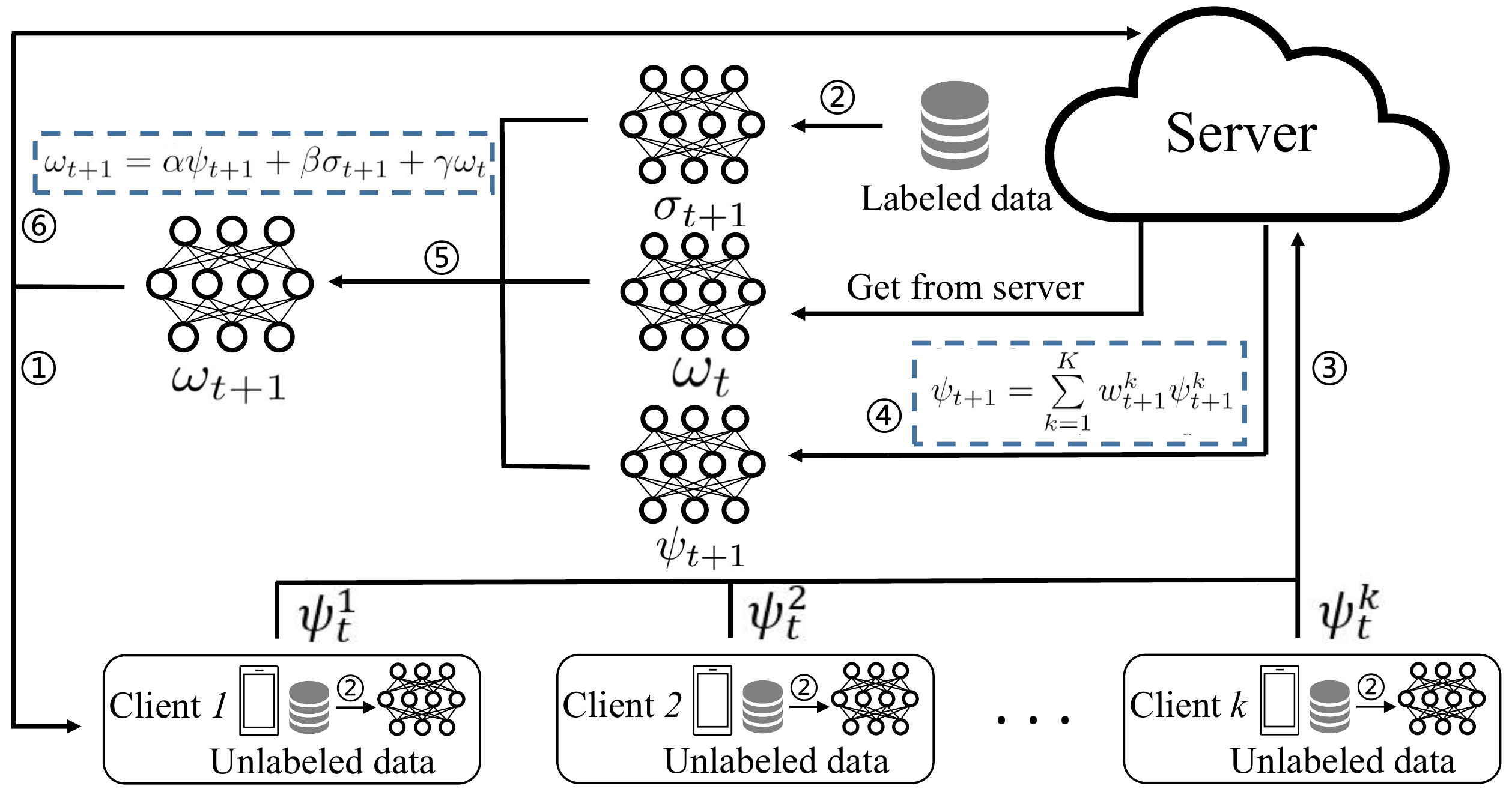}
	\caption{Overview of semi-supervised federated learning system.}
	\label{fig-1}
\end{figure*}

\section{Problem Definition}
In the SSFL system, there are two essential scenarios of SSFL based on the location of the labeled data. The first scenario considers a conventional case where clients have both labeled and unlabeled data (\emph{labels-at-client}), and the second scenario considers a more challenging case, where the labeled data is only available at the server (\emph{labels-at-server}). In particular, in this paper, we consider only the labels-at-server scenario. Next, we give the definition of the problem studied in this paper as follows:

\textbf{Labels-at-server Scenario}: In SSFL, we assume that there is a server $\mathcal{S}$ and $K$ clients, where the server holds a labeled dataset $\mathcal{D}_s=\{(x_i, y_i)\}_{i=1} ^{n}$ and each client holds a local unlabeled dataset $\mathcal{D}_k=\{u_i\}_{i=1} ^{m}$. Thus, in this scenario, for unlabeled training sample $u_i$, let $\mathcal{L}_u^k$ be the loss function at the client side:
\begin{equation}
\begin{aligned}
{\cal L}_u^k &= \frac{1}{m}\sum\nolimits_{{u_i} \in {{\cal D}_k}} C E(\widehat {{y_i}},{f_{{\theta _k}}}({u_i}))\\
& + \frac{1}{m}\sum\nolimits_{{u_i} \in {{\cal D}_k}} | |{f_{{\theta _k}}}({u_i}) - {f_{{\theta _k}}}(\pi ({u_i}))|{|^2},  
\end{aligned}
\end{equation}
where $m$ is the number of unlabeled samples, $\pi(\cdot)$ is the data augmentation function (e.g., flip and shift of the unlabeled samples), $\hat{y_i}$ is the pseudo label of unlabeled sample $u_i$, and $f_{\theta_k} (u_i)$ indicates the output of unlabeled sample $u_i$ on model $\theta_k$ of the $k$-th client.
For labeled sample $x_i$, let $\mathcal{L}_s$ be the loss function at the server side:
\begin{equation}
{{\cal L}_s} = \frac{1}{n}\sum\nolimits_{{x_i},{y_i} \in {{\cal D}_s}} C E({y_i},{f_\theta }({x_i})),
\end{equation}
where $n$ is the number of labeled samples, and $f_\theta (x_i)$ indicates the output of labeled sample $x_i$ on model $\theta$.

Therefore, the objective function of this scenario in SSFL system is to minimize the following loss function: 
\begin{equation}\label{eq-5}
\min {\cal L},\text{where} \quad {\cal L} \buildrel\textstyle.\over= \sum\limits_{k = 1}^K {{\cal L}_u^k}  + {{\cal L}_s}.
\end{equation}

Note that the whole learning process is similar to the traditional FL system, except that the server not only aggregates the client model parameters but also trains the model with labeled data.

\section{Algorithm and System Design}
\subsection{Semi-supervised Federated Learning System Design}
In our system setting, the server $\mathcal{S}$ holds a labeled dataset $\mathcal{D}_s = \{x_i, y_i\}_{i=1} ^{n}$, where $n$ indicates the number of labeled samples.
For $K$ clients, we assume that $k$-th client holds a local unlabeled dataset $\mathcal{D}_k = \{u_i \}_{i=1} ^{m}$, where $m$ denotes the number of unlabeled samples in the local client.
Similar to the traditional FL system, the server and clients in the SSFL are cooperative to train a high-performance global model $\omega^*$. The goal of previous work is to optimize the objective function mentioned above, i.e., Equation \eqref{eq-5}. However, they ignore the implicit contribution between iterations of the global model, which results in the learned global model not being optimal. Inspired by the above facts, we propose an SSFL algorithm called $\mathrm{FedMix}$ that focuses on the implicit contributions between iterations of the global model in a fine-grained manner. We define the supervised model trained on the labeled dataset as $\sigma$, the unsupervised model trained on the unlabeled dataset as $\psi$, and the aggregated global model as $\omega$. Specifically, we design a strategy that assigns three weights $\alpha ,\beta ,$ and $\gamma$ to the unsupervised model $\psi$, supervised model $\sigma$, and the previous round of global model, respectively. The designed algorithm can capture the implicit relationship between each iteration of the global model in a fine-grained manner. Thus, the steps of our proposed $\mathrm{FedMix}$ algorithm are as follows:
\begin{itemize}
    \item \textbf{\textit{Step 1, Initialization:}} The server randomly selects a certain proportion of $F$ ($0<F<1$) clients from all local clients to send the initialized global model $\omega_{0} $.
    Note that the global model $\omega_{0} $ also remains on the server-side.
   
    \item \textbf{\textit{Step 2, Server Training:}}
    Unlike FL, in our SSFL system, the server not only aggregates the model uploaded by the clients, but also trains the supervised model $\sigma$ (i.e., $ \sigma_t \leftarrow {\omega _t}$) on the labeled dataset $\mathcal{D}_s$.
    Thus, the server uses the local optimizer on the labeled dataset $\mathcal{D}_s$ to train the supervised model $\sigma$.
    The minimization of the objective function is defined as follows:
\begin{equation}
\mathop {\min }\limits_{\sigma  \in {\mathbb{R}^d}} {{\cal L}_s}({\sigma _t}), \text{where}\quad{{\cal L}_s}({\sigma _t}) \buildrel\textstyle.\over= {\lambda _s}CE(y,{f_{{\sigma _t}}}(x)),
\end{equation}
where $\lambda_s$ is the hyperparameter, ${x}$ and ${y}$ are from labeled dataset $\mathcal{D}_s$, and $f_{\sigma_t}({x})$ means the output of labeled samples on supervised model $\sigma$ at $t$-th training round.
    
    \item \textbf{\textit{Step 3, Local Training:}}
    The $k$-th client utilizes the local unlabeled data to train the received global model $\omega_{t}$ (i.e., $ \psi_t^k \leftarrow {\omega _t}$) and then obtains the unsupervised model $\psi_{t+1} ^{k}$.
    Thus, we define the following objective function:
\begin{equation}
\begin{aligned}
\mathop {\min }\limits_{\psi  \in {\mathbb{R}^d}} {{\cal L}_u}(\psi _t^k), {{\cal L}_u}(\psi _t^k) \buildrel\textstyle.\over= {\lambda _2}||{f_{\psi _t^k}}({\pi _1}(u)) - {f_{\psi _t^k}}({\pi _2}(u))|{|^2}\\
 + {\lambda _1}CE(\widehat y,{f_{\psi _t^k}}(u)) + {\lambda _{L1}}||{\sigma _t} - \psi _t^k|{|^2},
\end{aligned}
\end{equation}
where $\lambda_1, \lambda_2,$ and $\lambda_{L1}$ are hyperparameters to control the ratio between the loss terms, $\psi_{t}^{k}$ is the unsupervised model of the $k$-th client at $t$-th training round, ${u}$ is from unlabeled dataset $\mathcal{D}_k$, $\pi (\cdot)$ is the form of perturbation, i.e., $\pi_1$ is the shift augmentation, $\pi_2$ is the flip augmentation, $||\sigma_{t} - \psi_{t}^{k}||^2$ is a penalty term that aims to let the $k$-th client unsupervised model $\psi_{t}^{k}$ learn the knowledge of the supervised model $\sigma_t$ (note that $\sigma_t$ is inferred from Equation (9)), and $\hat{y}$ is pseudo label obtained by using our proposed argmax method. The argmax method is defined as follows:
\begin{equation}
\hat{y} = \mathsf{\textbf{1}} (\mathrm{Max} (\sum\limits_{i=1} ^{A} f_{\psi_{t}^{k}}(\pi_i (u)))),
\end{equation}
where $\mathrm{Max} (\cdot)$ is a function that can output the maximum probability that unlabeled data belongs to a certain class, $\mathsf{\textbf{1}} (\cdot)$ is the one-hot function that can change the numerical value to 1, $A$ represents the number of unlabeled data after data augmentation, and ${u}$ is from unlabeled dataset $\mathcal{D}_k$. Specifically, we discard low-confident predictions below confidence threshold $\tau =0.80$ when generating pseudo-labels.
\begin{figure*}[!t]
	\centering
	\includegraphics[width=0.8\linewidth]{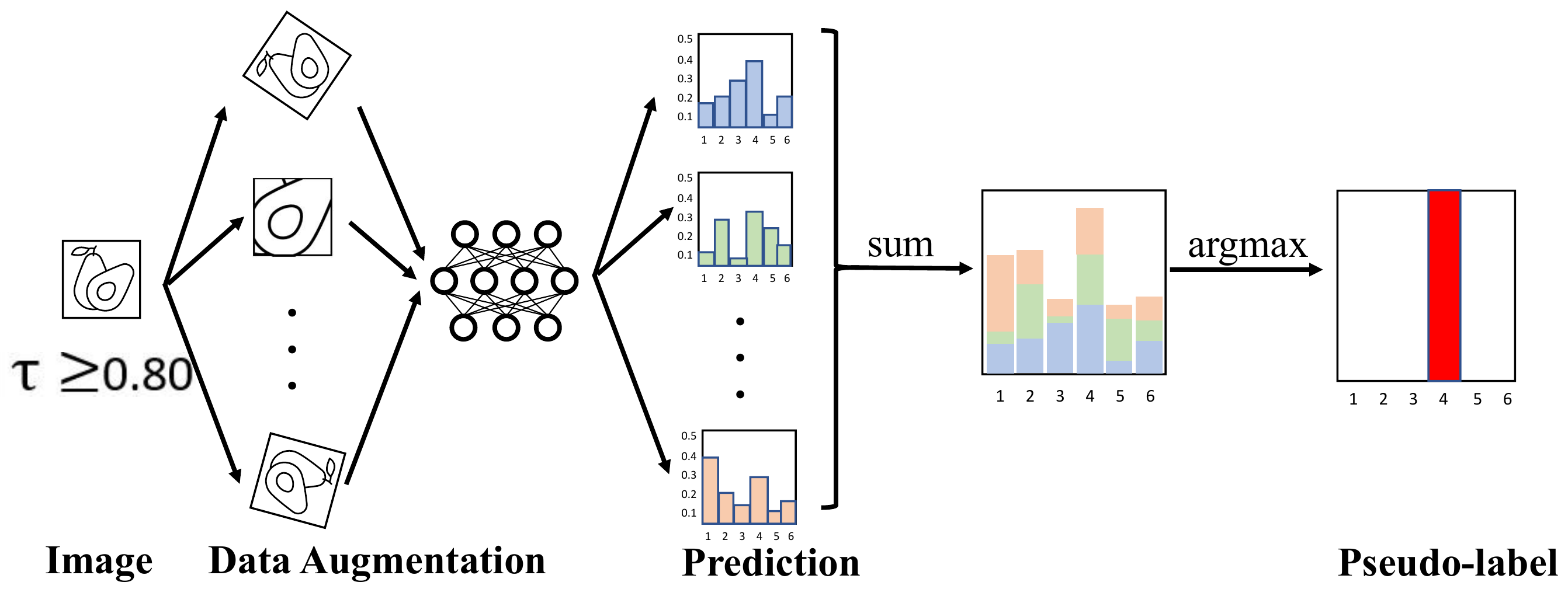}
	\caption{Overview of the proposed argmax method.}
	\label{fig-2}
\end{figure*}
\addtolength{\topmargin}{0.1cm}
\item \textbf{\textit {Step 4, Aggregation:}}
The server uses the proposed $\mathrm{FedFreq}$ (see Section IV-B) aggregation algorithm to aggregate the unsupervised models uploaded by the clients to obatin the global unsupervised model, i.e.,$\psi_{t+1} = \sum\limits_{k=1} ^{K} w_{t+1}^{k} \psi_{t+1} ^{k},$ where $\psi_{t+1} ^{k}$ is the unsupervised model of the $k$-th client at $t+1$-th training round and $w_{t+1}^{k}$ is the weight of the $k$-th client. The server then aggregates the global unsupervised model $\psi_{t+1}$, the supervised model $\sigma_{t+1}$, and the global model $\omega_{t}$ from the previous round $t$ to obtain a new global model $\omega_{t+1}$:
\begin{equation}
\omega_{t+1} = \alpha \psi_{t+1} + \beta \sigma_{t+1} + \gamma \omega_{t},
\end{equation}
where $\alpha$, $\beta$, and $\gamma$ are the corresponding weights of the three models and $(\alpha ,\beta ,\gamma ) \in \{ \alpha  + \beta  + \gamma  = 1 \wedge \alpha ,\beta ,\gamma  \geqslant 0\}$.
\end{itemize}

Repeat all the above steps until the global model converges. The proposed $\mathrm{FedMix}$ algorithm is shown in Algorithm 1.

\begin{algorithm}[!t]\label{algorithm}
	\caption{$\mathrm{FedMix}$ algorithm on labels-at-server scenario.}
	\begin{algorithmic}[1]
	\REQUIRE The client set $\mathcal{K}$, $B_{server}$ is the mini-batch size at the server side, $E_{server}$ is the number of epochs at the server side,
		$B_{client}$ is the local mini-batch size at the client side,
		$E_{client}$ is the number of local epochs at the client side,
		and $\eta$ is the learning rate.
	\ENSURE The optimal global model $\omega^\ast$.
	\STATE \textbf{Server executes:}
	\STATE Initialize global model $\omega_0$
	\FOR{each round $t = 0, 1, 2, ...$}
	\STATE $\sigma_{t}$ $\gets$ $\omega_{t}$
		  \FOR{the server epoch $e$ from $1$ to $E_{server}$}
		    \FOR{mini-batch $b \in B_{server}$}
		    \STATE $\sigma_{t+1}$ = $\sigma_t$ - $\eta\bigtriangledown\mathcal{L}_{s}(\sigma_t, \mathcal{D}_s, b)$\\
		    \ENDFOR
		  \ENDFOR
		  \STATE $m \gets \mathrm{max}(F\cdot K, 1)$\\
		\STATE $S_t \gets$ randomly select $m$ clients from the client set $\mathcal{K}$\\
		  \FOR{each client $k \in S_t$ $\textbf{in parallel}$}
		  \STATE $\psi_{t}^k$ $\gets$ $\omega_{t}$
		  \STATE $\psi_{t+1}^{k} \gets$ \textbf{ClientUpdate($k, \psi_{t}^{k}$)}\\
		  \ENDFOR
		\STATE $\psi_{t+1} = \sum\limits_{k=1} ^{K} w_{t+1}^{k} \psi_{t+1} ^{k}$  $// $ Refer to $\mathrm{FedFreq}$ algorithm\\
		\STATE$\omega_{t+1} = \alpha \psi_{t+1} + \beta \sigma_{t+1} + \gamma \omega_{t}$\\
		\ENDFOR
		\STATE \textbf{ClientUpdate($k$, $\psi_{t}^{k}$):}$//$ Run on client $k$\\
		\FOR{each local epoch $e$ from $1$ to $E_{client}$}
		    \FOR{minibatch $b \in B_{client}$}
		    \STATE$\psi_{t+1}^{k}$ = $\psi_t^{k}$ - $\eta\bigtriangledown\mathcal{L}_{u}(k, \psi_{t}^{k}, \mathcal{D}_k, b)$\\
		    \ENDFOR
		  \ENDFOR
			\RETURN $\omega^\ast$ to server.
		\end{algorithmic}
	\end{algorithm}

\subsection{$\mathrm{FedFreq}$ Aggregation Algorithm}
In this section, we present the designed $\mathrm{FedFreq}$ aggregation algorithm, which can dynamically adjust the weight of the corresponding local model according to the training frequency of the client to alleviate the non-IID problem. We observe that the parameter distribution of the global model will be biased towards clients that often participate in federated training, which is obviously not friendly to the robustness of the global model. Therefore, our insight is to reduce the influence of clients with high training frequency on the global model to improve the robustness of the model. Thus, the formal expression of the $\mathrm{FedFreq}$ aggregation algorithm is as follows:
\begin{equation}
\begin{aligned}
w_{t+1}^{k} = \dfrac{1 - p_{t+1}^{k}} 
{1 - p_{t+1}^{1}+ \cdots +1-p_{t+1}^{k}}
=\dfrac{1 - p_{t+1}^{k}}{FK-1},
\end{aligned}
\end{equation}
where $F$ is the sample proportion of the server, $K$ is the total number of clients, $p_{t+1}^{k} = \dfrac{q_{t+1}^{k}}{\sum\nolimits_{k \in S_{t+1}}q_{t+1}^{k}}$, $q_{t+1}^{k}$ is the number of times that the $k$-th client has been trained up to the $t+1$-th round, and $S_{t+1}$ denotes the set of clients selected by the server in round $t+1$. Then, for the client, the $\mathrm{FedFreq}$ aggregation rule is expressed as follows: $\psi_{t+1} = \sum\limits_{k=1} ^{K} w_{t+1}^{k} \psi_{t+1} ^{k}.$

\subsection{Dirchlet Data Distribution Function}
\begin{figure}[t]
	\centering
	\subfigure []{\includegraphics[width=0.6\linewidth]{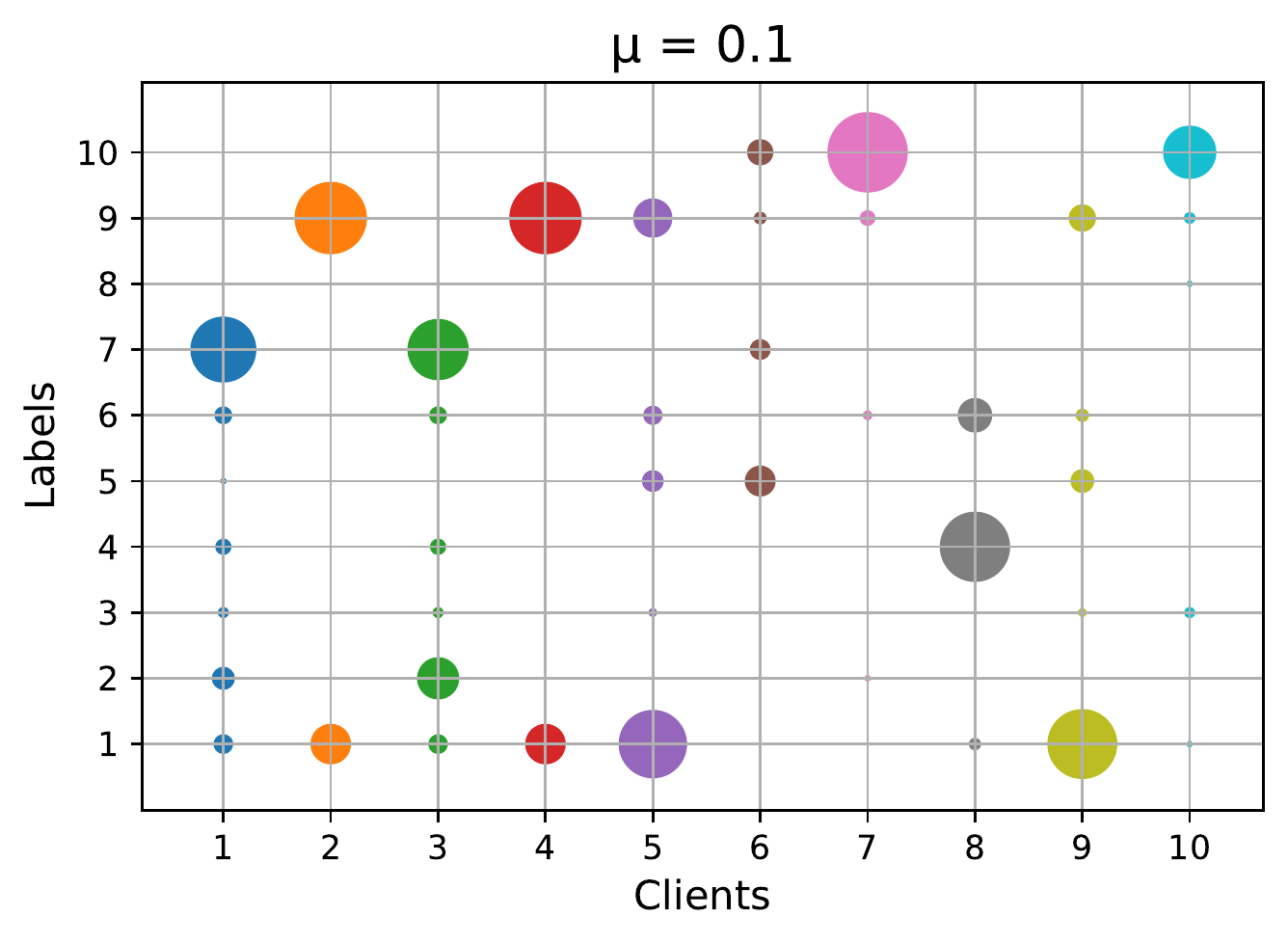}
		\label{c-11}}
	\subfigure[]{	\includegraphics[width=0.6\linewidth]{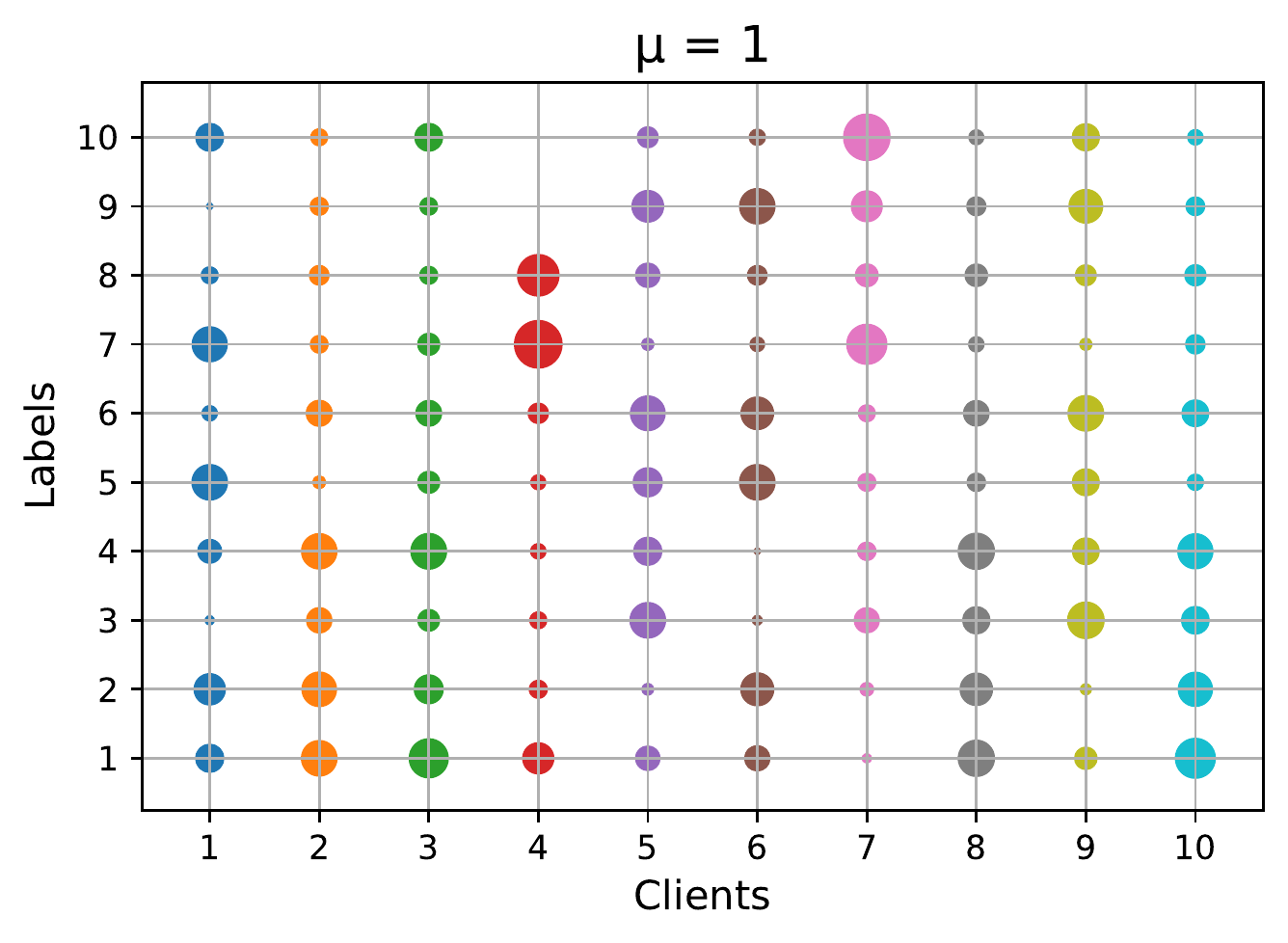}
		\label{c-21}}
 	\subfigure[]{	\includegraphics[width=0.6\linewidth]{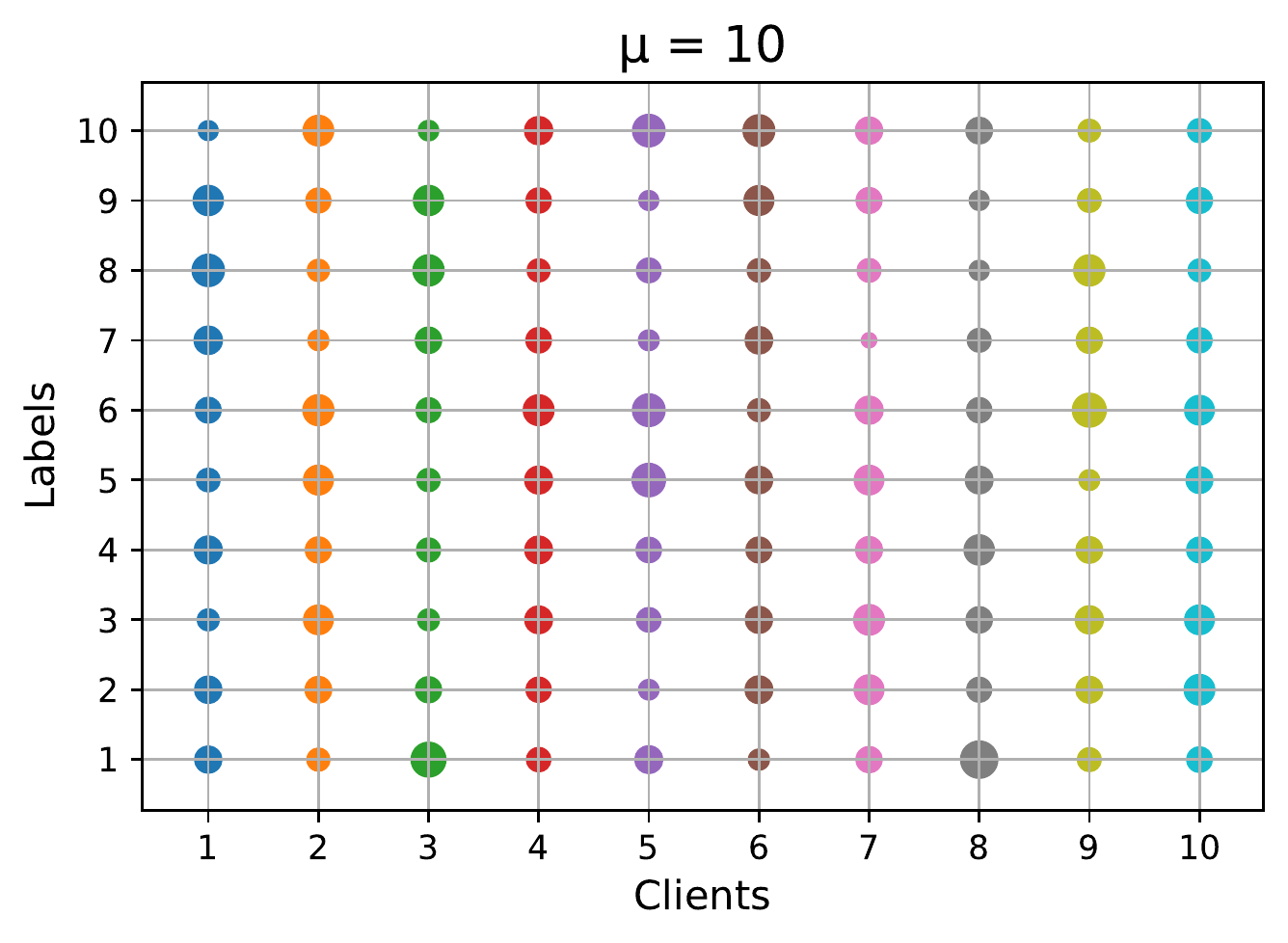}
 		\label{c-31}}
	\caption{Overview of the data distribution when $\mu =\{0.1,1,10\}$.}
	\label{fig-3}
\end{figure}

To better evaluate the robustness of the designed system to non-IID data, in this paper, we introduce the Dirchlet distribution function \cite{yurochkin2019bayesian,hsu2019measuring}, which is a popular non-IID function, to adjust the non-IID level of the local client data. Specifically, we generate data distributions of different non-IID levels by adjusting the parameter (i.e., $\mu$) of the Dirchlet distribution function. We assume that the local dataset $\mathcal{D}_k$ of $k$-th client has $c$ classes, and thus, the definition of Dirichlet distribution function is as follows:
\begin{equation}
P_{k} (\varphi_{1}, ..., \varphi_{c}) = \dfrac{\Gamma(\sum\nolimits_{i} \mu_i)}{\prod_{i}\Gamma(\mu_i)} \prod_{i=1}^{c}\Gamma(\mu_i) \varphi_{i}^{\mu_{i} -1},
\end{equation}

\begin{equation}
p_{k}(\varphi_{c}) = \dfrac{P_{k}(\varphi_{c})}{\sum\limits_{i=1}^{c} P_{k}(\varphi_{i})},
\end{equation}
where $\Theta$ is a set of $c$ samples randomly selected from the Dirichlet function, i.e., $\Theta  = \{ {\varphi _1}, \ldots ,{\varphi _c}\} $ and $\Theta \sim Dir(\mu_{1}, \cdots,\mu_{c})$, $\mu, \mu_1, .., \mu_c$ are the parameters of the Dirichlet distribution function (where $\mu=\mu_1=\mu_2=...=\mu_c$), and $p_{k} (\varphi_c)$ denotes the proportion of the $c$-th class data in all data of the client. In particular, the smaller the $\mu $, the higher the non-IID level of the data distribution of each client; otherwise, the data distribution of the client tends to the IID setting. Therefore, we adjust the parameters of the Dirichlet distribution function to simulate the different non-IID levels of the client's local dataset. For example, as shown in Fig. \ref{fig-3}, we demonstrate the data distribution when $\mu =\{0.1,1,10\}$.

\section{Experiment}
In the Labels-at-Server scenario, we compared our method $\mathrm{FedMix}$ and the baseline $\mathrm{FedMatch}$ \cite{jeong2020federated} on the CIFAR-10 dataset. Furthermore, we simulate the federated learning setup (one server and $K$ clients) on a commodity machine with Intel(R) Core(TM) i9-9900K CPU @ 3.60GHz and NVIDIA GeForce RTX 2080Ti GPU.
\subsection{Experiment Setup}
\textbf{Dataset 1) CIFAR-10 dataset under IID setting:} We use the CIFAR-10 dataset including 56,000 training samples and 2,000 test samples as the validation dataset in our experiment. The training set includes 55,000 unlabeled samples and 1,000 labeled samples, where the former is used to train the unsupervised model at the local and the latter is used to train the supervised model at the server. The unlabeled samples are equally distributed to 100 clients at a ratio of $1:100$, of which there are 550 samples for each client (i.e., 55 for each class, and 10 classes in total). Similarly, the labeled samples have a total of 1,000 and contain 10 classes on the server, of which there are 100 samples in each class. Meanwhile, we set a participation rate $F=0.05$ of clients, i.e., 5 clients are randomly selected for training in each round.

\textbf{Dataset 2) CIFAR-10 dataset under non-IID setting:} Our setting is similar to the above IID setting, except that the Dirchlet distribution function is introduced to adjust the non-IID level of the local client data. Specifically, we generate data distributions of different non-IID levels by adjusting the parameter (i.e., $ \mu $) of the Dirchlet distribution function. Meanwhile, we simulate quantity imbalance and class imbalance the local client data. In particular, we make each client hold a different number of training samples. For example, some customers have 580 samples, while some customers have less than 50 samples. Second, we make the client hold a different sample size for each category in the data. For example, some clients have ten types of data, and some clients have less than two types of data.



\textbf{Baseline and training details:} Our baseline is $\mathrm{FedMatch}$ \cite{jeong2020federated} naively using unsupervised model and supervised model parameter decomposition strategy, i.e., $\omega = \psi + \sigma$.
In the training process, both our model and baseline use Stochastic Gradient Descent (SGD) to optimize the ResNet-9 neural network with initial learning rate $\eta=1e-3$. We set training round $t=150$, the number of labeled samples on sever is $N_s=1000$, local client training epoch $E_{client}=1$ and mini-batch size $B_{client}=64$, the server training epoch $E_{server}=1$ and mini-batch size $B_{server}=64$. Second, we set the data augmentation number in the argmax method $A = 5$.
\subsection{Experiment Results}
\begin{figure}[!t]
	\centering
	\large
	\subfigure []{\includegraphics[width=0.45\linewidth]{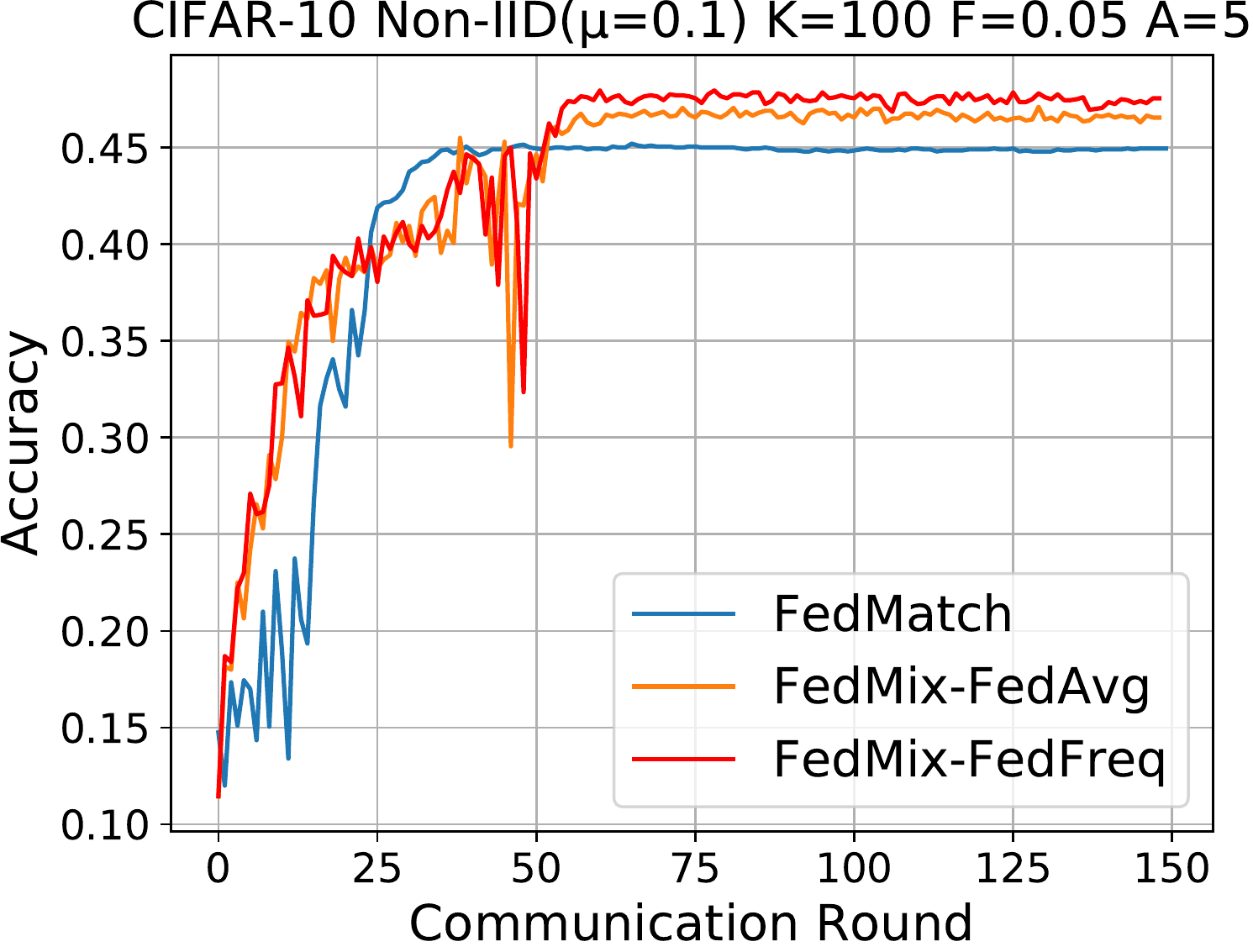}
		\label{a-11}}
	\hfill
	\subfigure[]{	\includegraphics[width=0.45\linewidth]{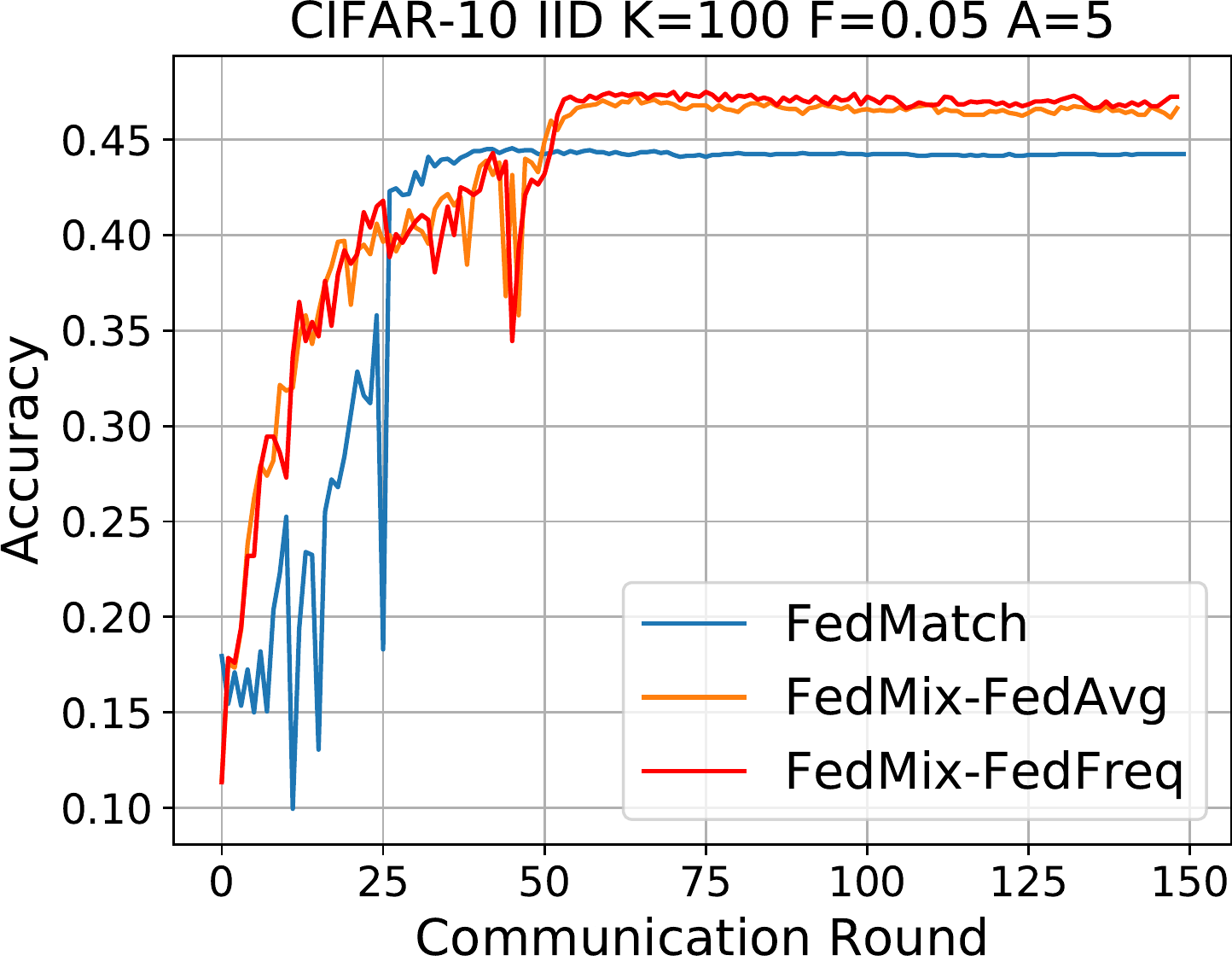}
		\label{b-11}}
	\caption{Test accuracy curves on IID and non-IID setting.}
	\label{fig-4}
\end{figure}

As shown in Fig. \ref{fig-4}, under IID and non-IID settings, our method $\mathrm{FedMix}$ is better than baseline under each different aggregation method settings. For example, under a non-IID setting, the convergence accuracy of our method is 47.5\% about 3\% higher than that of the baseline. In particular, the accuracy of our method increases faster and more stable in the early stage of model training. The reason is that: (1) The $\mathrm{FedMix}$ focuses on the implicit contributions between iterations of the global model in a fine-grained manner, while the $\mathrm{FedMatch}$ only naively uses model parameter decomposition. (2) Frequency-based aggregation method $\mathrm{FedFreq}$ is more suitable for non-IID settings.
Notably, the $\mathrm{FedFreq}$ only requires the server to give appropriate weights in the aggregation process according to the training frequency of each client, which does not bring additional computational overhead to the server and local clients.

\begin{figure}[!t]
	\centering
	\large
	\subfigure []{\includegraphics[width=0.45\linewidth]{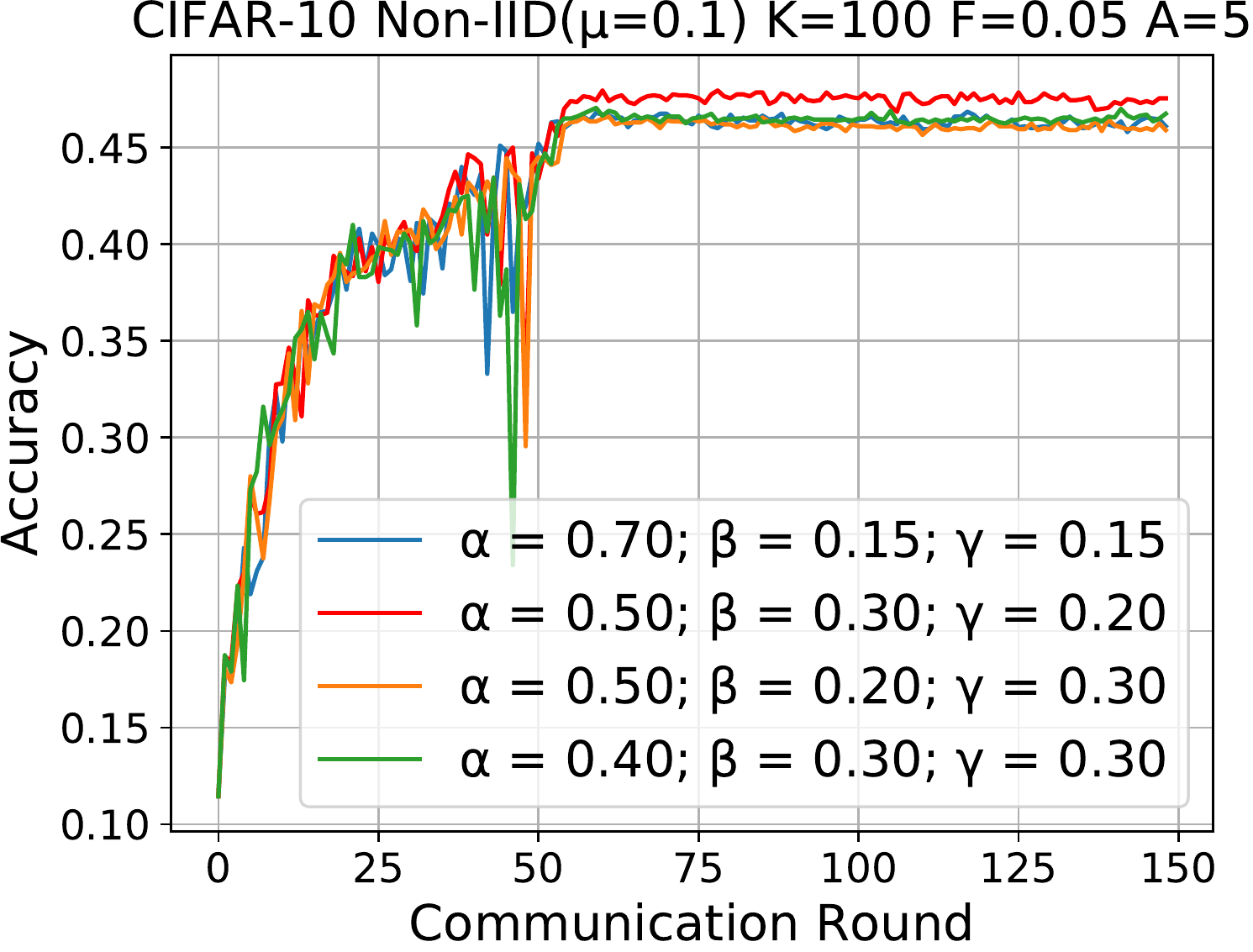}
		\label{a-21}}
	\hfill
	\subfigure[]{	\includegraphics[width=0.45\linewidth]{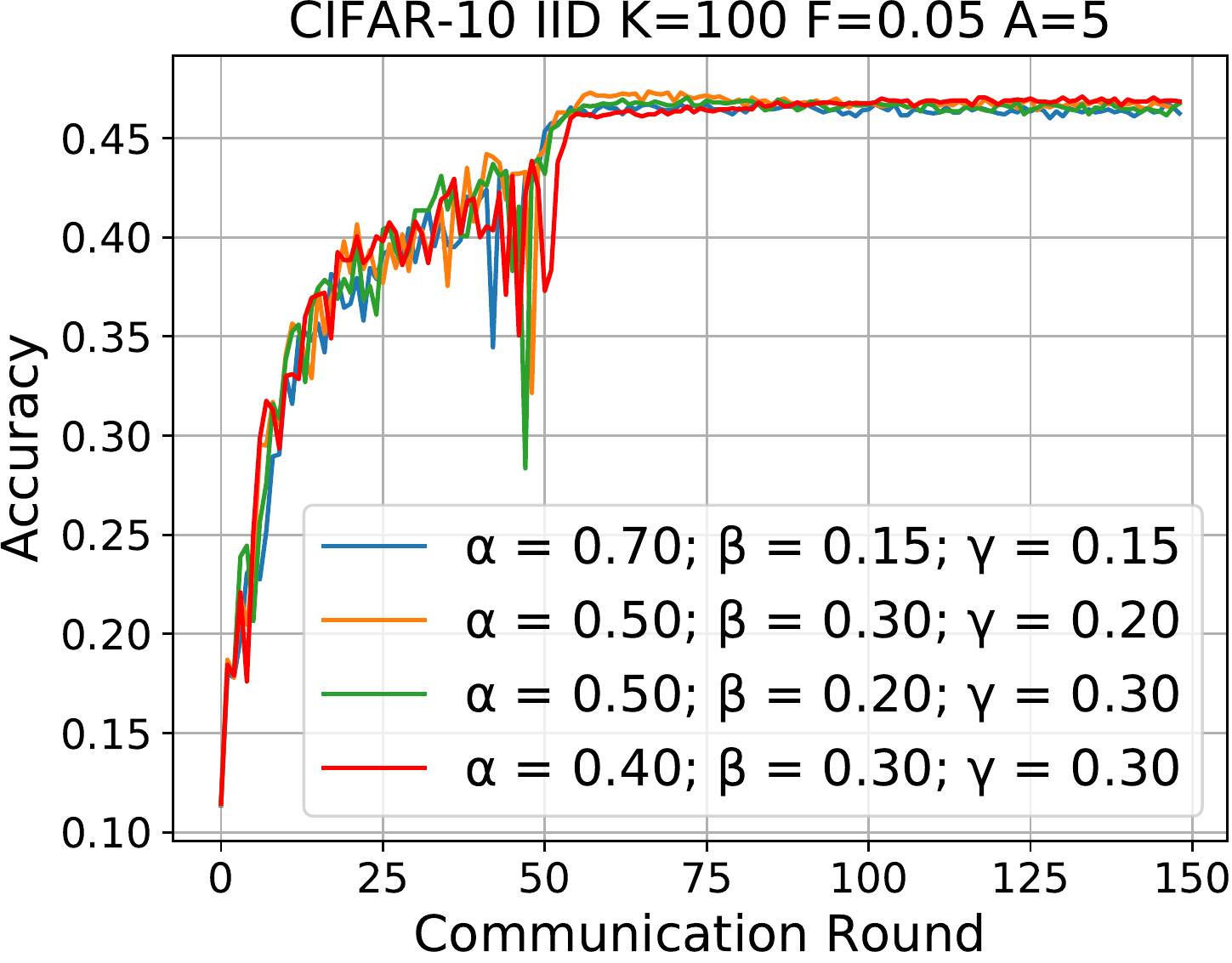}
		\label{b-22}}
	\caption{Test accuracy curves under different hyperparameter settings.}
	\label{fig-5}
\end{figure}

\begin{figure}[!t]
	\centering
	\large
	\subfigure []{\includegraphics[width=0.45\linewidth]{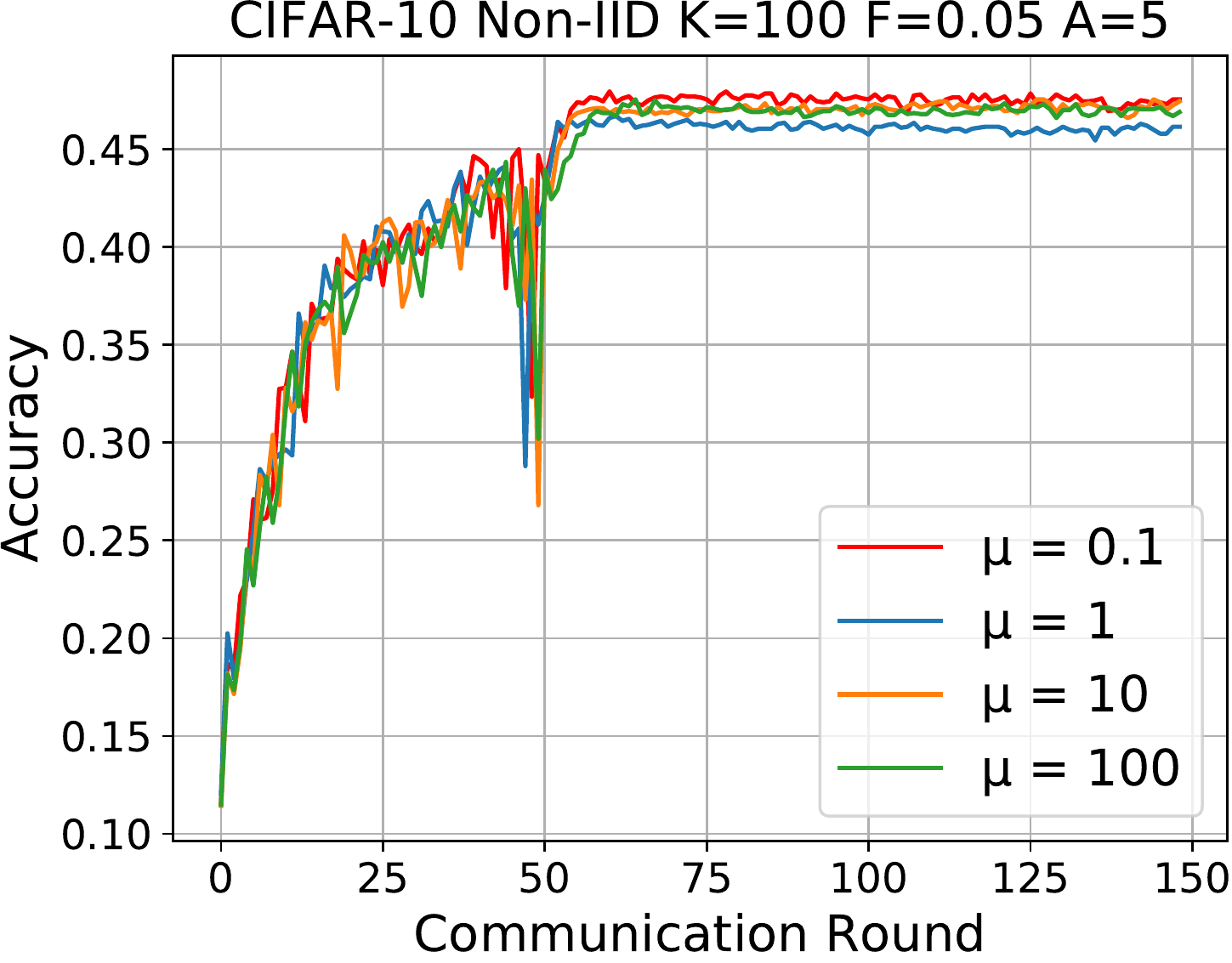}
		\label{a-31}}
	\hfill
	\subfigure[]{	\includegraphics[width=0.45\linewidth]{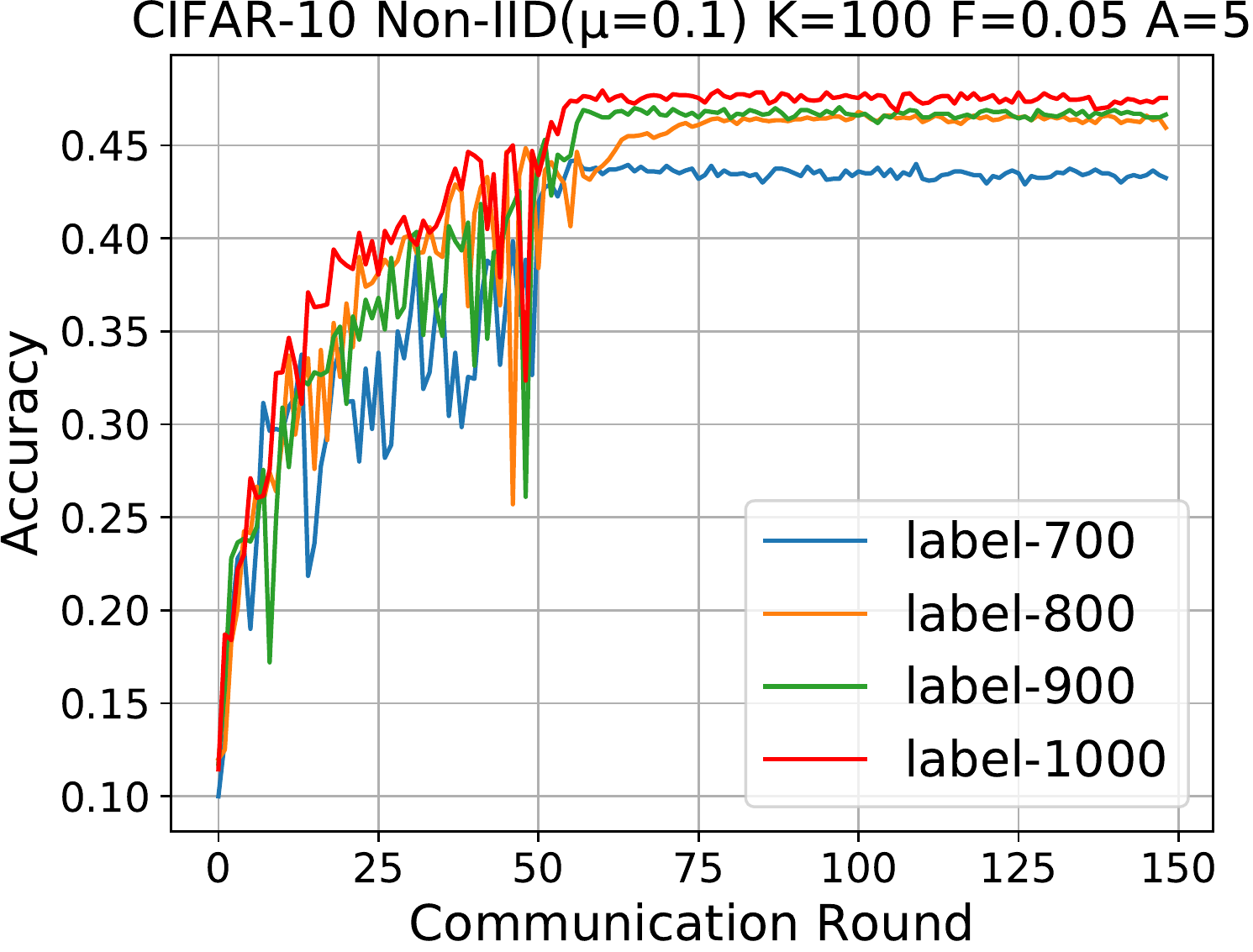}
		\label{b-32}}
	\caption{(a) Performance comparison on different non-IID levels; (b) Performance comparison on different numbers of labeled samples.}
	\label{fig-6}
\end{figure}

Fig. \ref{fig-5} shows the performance comparison of the proposed method under different hyperparameter settings. To be specific, the three hyperparameters are the weights of the global unsupervised model, the supervised model, and the previous round of global model.
From Fig. \ref{fig-5}, we can find that under the non-IID setting, as $\alpha$ decreases, the accuracy curve of the proposed method becomes unstable.
The reason for this phenomenon is that: with the decrease of global unsupervised model weight, $\mathrm{FedFreq}$ is losing its effect.
In particular, when $\alpha=0.5$, $\beta=0.3$, and $\gamma=0.2$, our global model is the aggregation of the optimal weights of the three models. In addition, we find that the proposed method is easy to achieve better performance when these three parameters are relatively close under the IID setting, as shown in Fig. \ref{b-22}.


Fig. \ref{a-31} shows the performance comparison of the proposed method on different non-IID levels of client data. In this experiment, we let $\mu=0.1$ denote the highest non-IID level of the client data. In this case, as the value of $\mu$ increases, the local client data distribution is closer to the IID setting. 
It can be seen from Fig. \ref{fig-6} that for different non-IID levels, our method all can achieve stable accuracy. Meanwhile, the model convergence accuracy under $\mu =\{ 0.1, 1, 10, 100\}$ settings does not differ by more than 1\%. Therefore, our method is not sensitive to the different levels of client data distribution, i.e., it is robust to different types of data distribution settings.

Fig. \ref{b-32} shows the performance comparison of the proposed method in the case of different numbers of labeled samples at the server. Obviously, the converged accuracy of our method is 47\% with 800 labeled samples, which is 2\% higher than $\mathrm{FedMatch}$. 
However, when the number of labeled samples is reduced to 700, the accuracy of our model decreases greatly. Therefore, we regard $N_s=800$ as the best setting for our method.
	\vspace{-0.1cm}
\subsection{Analysis}
In this section, we further analyze the advantages of $\mathrm{FedMix}$ compared to $\mathrm{FedMatch}$ in labels-at-server scenario.

\textbf{1) The performance of $\mathrm{FedMix}$ training model under the CIFAR-10 dataset is better than $\mathrm{FedMatch}$.} This is due to $\mathrm{FedMatch}$ simply uses the strategy of parameter decomposition of unsupervised model and supervised model in the training process, i.e., $\omega_t = \psi_t + \sigma_t$. In this way, the learned global model will be biased towards unlabeled data (unsupervised model) or labeled data (supervised model) instead of the overall data.
Thus, in order to avoid the drift problem of the global model, $\mathrm{FedMix}$ adds the global model from the previous round to the model parameter aggregation, i.e., $\omega_{t} = \alpha \psi_{t} + \beta \sigma_{t} + \gamma \omega_{t-1}$.
Meanwhile, we conducted a sensitivity experiment of model performance to different hyperparameter weights to find the optimal weight combination.

\textbf{2) $\mathrm{FedMix}$ is robust to different levels of non-IID data.} In our experiment, we introduced the Dirichlet distribution function to simulate the local client non-IID data in FL. In details, we generate data distributions of different non-IID levels by adjusting the parameters of the Dirchlet distribution function, i.e., $\mu =\{ 0.1, 1, 10, 100\}$ respectively correspond to different levels of non-IID. The results show that the performance difference of our model does not exceed 1\% under different levels of non-IID settings.
$\mathrm{FedMatch}$ uses a pseudo-random method to generate the non-IID data distribution of each client. However, in reality there is no such distribution, which will cause the model to lose its robustness.
\section{conclusion}
In this paper, we studied the labels-at-server scenario and addressed the problem of data availability and non-IID in FL. To solve the first problem, we designed a robust SSFL system that uses the $\mathrm{FedMix}$ algorithm to achieve high-precision semi-supervised learning. To tackle the non-IID problem, we propose a novel aggregation algorithm $\mathrm{FedFreq}$, which effectively achieves the stable performance of the global model in the training process without adding additional computational overhead.
Through experimental verification, our robust SSFL system is significantly better than the baseline in performance. In future work, we will further improve the algorithm to maximize the use of unlabeled data. Furthermore, we will continue to strengthen the theory of SSFL so that it can be better applied in real-world scenarios.

\section*{Acknowledgment}
This work is supported by Heilongjiang Provincial Natural Science Foundation of China (Grant No. LH2020F044), the 2019--``Chunhui Plan'' Cooperative Scientific Research Project of Ministry of Education of China (Grant No. HLJ2019015), the Fundamental Research Funds for Heilongjiang Universities, China (Grant No. 2020-KYYWF-1014), and also supported by NSFC under grant No. 6210071210.
\bibliographystyle{IEEEtran}
\bibliography{ref}
\end{document}